\documentclass[conference,onecolumn]{IEEEtran}

\usepackage[ruled]{algorithm}
\usepackage{algorithmic}
\usepackage[pdftex]{graphicx}
\usepackage{caption}
\usepackage{caption3}

\usepackage{stfloats}
\usepackage[sort, numbers]{natbib}
\usepackage{mathrsfs}
\usepackage{algorithm}
\usepackage{graphics}
\usepackage{multirow}
\usepackage{parskip}
\usepackage{array}
\newcolumntype{P}[1]{>{\centering\arraybackslash}p{#1}}
\usepackage{amsmath}

\ifCLASSINFOpdf
  
\else
 
\fi

\begin{document}

\title{CRDT: Correlation Ratio Based Decision Tree Model for Healthcare Data Mining}

\author{\IEEEauthorblockN{Smita Roy}
\IEEEauthorblockA{Department of Computer Science\\Central University of Bihar\\
Patna - 800 014, Bihar, India,\\
Email: {smitaroy}@cub.ac.in}
\and
\IEEEauthorblockN{Samrat Mondal and Asif Ekbal}
\IEEEauthorblockA{Department of Computer Science and Engineering\\Indian Institute of Technology Patna\\
Patna - 800 013,Bihar,India\\
Email: {samrat,asif}@iitp.ac.in}

}

\maketitle

\begin{abstract}
  The phenomenal growth in the healthcare data has inspired us in investigating robust and scalable models for data mining. For classification problems Information Gain(IG) based Decision Tree is one of the popular choices. However, depending upon the nature of the dataset, IG based Decision Tree may not always perform well as it prefers the attribute with more number of distinct values as the splitting attribute. Healthcare datasets generally have many attributes and each attribute generally has many distinct values. In this paper, we have tried to focus on this characteristics of the datasets while analysing the performance of our proposed approach which is a variant of Decision Tree model and uses the concept of Correlation Ratio(CR). Unlike IG based approach, this CR based approach has no biasness towards the attribute with more number of distinct values. We have applied our model on some benchmark healthcare datasets to show the effectiveness of the proposed technique.
\end{abstract}


 \section{Introduction}\label{secIntro}

Due to the growth of Internet technology and healthcare software, data are available in abundance in
unstructured form as all as structured form over the Internet. We
are rich in information but lack of knowledge. So this led to the
path to healthcare data mining. Healthcare data mining techniques
find hidden but useful patterns from healthcare datasets. 

Diseases like heart-diseases, hepatitis, diabetes are very common among all ages of
patients. Some alarming statistics regarding those diseases are given below:

\begin{itemize}
\item One-third of all deaths in India will be caused by cardiovascular disease by the year 2020.\cite{stat}
\end{itemize}

\begin{itemize}
\item In middle-income countries, diabetes is one among the top 10 life threatening diseases.\cite{Motamedi01}.
\end{itemize}

All these observations are made after some careful analysis of different healthcare datasets. And such observations often change the focus of the government policies. 

There are different data mining algorithms - such as Decision Tree
algorithm\cite{Han06}, Naive Bayes Classifier \cite{George01}\cite{Domingos01}\cite{Han06}, Neural Network model \cite{Murata01}\cite{Han06}, k-Nearest Neighbour algorithm\cite{Han06}, Support Vector Machine \cite{Vapnik01}\cite{Han06}, K-means \cite{MacQeen01}\cite{Han06}, Bisecting K-means algorithm \cite{Steinbach01}, Association Rule mining algorithms \cite{Agrawal01}\cite{Han06} etc. Among them some are used for classification, some for clustering purposes and some others for finding easily interpretable rules for taking proper decision. Decision Tree algorithm is a very popular algorithm for classifications. It generally  uses Information Gain (IG) as the criterion for splitting on an attribute. The attribute with the highest IG is chosen as the splitting attribute at each level. But Information Gain has some disadvantages like it prefers the attribute which has large number of distinct values. So if there is an attribute in the dataset like product-ID, then Information Gain approach will prefer to split on product-ID  as because this attribute can uniquely identify each tuple in the set and this would result in a large number of partitions (as many as there are values), each one will have just one tuple. Since there will be no records with different class labels in each partition, the required  information to classify data set D based on this partitioning based on Information Gain \cite{Han06} principle would be Info product-ID(D)=0. Therefore, the information gained by partitioning on this attribute is highest. Therefore, such a partitioning is useless for classification \cite{Han06}. So it will not  generalize the model. So, IG based approach is not effective for all types of datasets. For instance, if a dataset has different attributes with different numbers of distinct values, it prefers the attributes with more numbers of distinct values as splitting attributes though some other attributes with less number of distinct values may be more significant for classification. 
  
  Due to this reason for some dataset IG based approach does not provide adequate accuracy. To overcome this we have proposed an approach which uses the concept of Correlation Ratio \cite{cor_ratio2} or CR as the splitting criterion. This method has no such biasness. It considers that attribute for classification which is significant enough to identify at least one outcome class. The general CR method is suitable for quantative data. But our proposed CR based approach is applicable to nominal or categorical attributes also.

The organization of the paper is as follows: Section \ref{secRW}
describes the related work done in this area. Section
\ref{secProp} illustrates our proposed approach. The result and analysis of the proposed approach is shown in Section \ref{sectabdata}. Section \ref{secConclusion} concludes the paper and gives future direction of the work.

\section{Related Work}\label{secRW}
Decision Tree technique has been popularly used for finding interesting patterns in health care datasets. We will discuss next some of the relevant works demonstrating this fact.

Polat et al.\cite{Polat01} proposed a hybrid model for classification of multiclass dataset. For each class separate models based on C4.5 Decision Tree algorithm has been constructed and the class for which the model is built is given positive class label and the rest of the classes are assigned negative class label. The proposed model showed significant performance improvement over the traditional C4.5 Decision Tree based model. As optimization of dataset can improve classification accuracy, some more methods like Homogeneity-based algorithm (HBA) etc have been proposed by Pham et al.\cite{Huy05}. This algorithm was used in association with standard classification algorithms as SVM, DT and ANN to enhance their performances. The four parameters of HBA were then optimized by Genetic Algorithm. The proposed approach showed significant performance improvement over standard approaches. Decision Tree induction method has a wide variety of applications as discussed previously. Changala et al. \cite{Changala01} have discussed different aspects of the Decision Tree
induction method in the paper. Since most of the learning algorithms require the dataset to be in memory, it is a matter of concern for huge datasets. So, in this paper the scalability issues have also been discussed.

Karaolis et al.\cite{Minas01} used Decision Tree based models to find out the risk factors for three types of Coronary heart disease events - myocardial infarction (MI), percutaneous coronary intervention (PCI), and coronary artery bypass graft surgery (CABG). The models for PCI and CABG have performed well with 75\% classification accuracy compared to the model for MI. A predictive model for determining the ability of the persons affected with dementia to take help of a technology based on mobile phone based video streaming system has been developed by Zhang et al.\cite{Zhang01}. The dataset was having two classes : Adopter and Non-Adopter. Popular classification algorithms were used for building models. Experimental results shows that among all the models DT, NN, SVM and kNN based models performed well. G. Sathyadevi \cite{Sathyadevi11} proposed to build intelligent decision support system for Hepatitis disease diagnosis. At first most relevant attributes were selected based on some threshold value or some condition. Then CART(Classification And Regression Trees) Decision Tree algorithm was applied on the dataset which showed 83.2\% accuracy and it was relatively higher as compared to the accuracy obtained from models using ID3 and C4.5 Decision Tree algorithms. Some significant rules were extracted after constructing the Decision Tree using CART. Decision Tree as a prediction model has also been used by the author in \cite{Eldin01} to predict \emph{hepatitis C} virus(HCV) polyprotein cleavage sites. The challenge was to collect accurate data. The model gave a very good result with a 96\% accuracy which was slightly lesser than the
accuracy(97\%) of the model based on Support Vector Machine(SVM).

Overall the above survey demonstrates the effectiveness of Decision Tree technique for efficient healthcare.

\section{Proposed Approach}\label{secProp}
In this section, we give the details of our proposed Correlation Ratio based Decision Tree construction approach.

\subsection{Overview of Correlation Ratio}\label{subsecCRatio}

Sometimes, the expected outcome of our learning algorithm is some
categorical values like ``yes/no''. The Correlation Coefficient
method is suitable for applications where the outcome is
quantitative, thus it cannot be applied in those cases where
categorical outcome is desired. To sort out this problem, the
CR\cite{cor_ratio2} method can be applied.

\begin{table}[ht]
\captionsetup{justification=centering}
\caption{\textit{Some frequently used terms}}

\label{tab:13}
\centering
\resizebox{10cm}{!}{

\begin{tabular}{| P{2cm} | P{6cm} |}
\hline
\textbf{Notation} & \textbf{Corresponding Meaning}\\
\hline
$Y$ & \textit{Set of outcomes or class labels}\\
\hline
$\ell$ & \textit{Total number of samples} \\
\hline
$\ell_y$ & \textit{Number of times that outcome $y \in Y$ occurs is $\ell_y$}\\
\hline
$x_{jy}^{(i)}$ & \textit{$i$-th attribute value of the $j$-th tuple among the $\ell_y $ samples having outcome $y$}\\
\hline
$\overline{x}_y^{(i)}$ & \textit{Average of the $i$-th attribute from all sample vectors within each outcome class y}\\
\hline
$\overline{x}^{(i)}$ & \textit{Overall average of the $i$-th attribute}\\
\hline
\end{tabular}
}

\end{table}

The CR method can be employed to
partition the sample dataset into different categories
according to the observed outcome. A significant attribute is one
which can identify at least one outcome class where
the average value of the attribute and the average on all classes
are remarkably different, otherwise that attribute would not be
useful to identify any outcome. Table \ref{tab:13} provides a summary of the different notations used in our proposed approach and the corresponding meanings.

Suppose that there is a set of $\ell$ tuples in a dataset. Let
the number of times that outcome $y \in Y$ occurs is $\ell_y$, so
that the dataset can be partitioned by their
outcome as follows:

\begin{equation}
\forall y \in Y|
\{S_y=(x_{jy}^{(1)},\cdots,x_{jy}^{(n)})\};j=1,\cdots,\ell_y  
\end{equation}

where $S_y$ is the set of all tuples with outcome $y$ and $x_{jy}^{(i)}$ is the value of the $i$ -th attribute of the $j$-th tuple among all the
$\ell_y $ tuples with outcome $y$ . The average of the $i$-th attribute from all tuples
within each outcome class is given by:

\begin{equation}
\forall y\in Y|  \overline{x}_y^{(i)}=\frac{\sum_{j=1}^{\ell_y}{x_{jy}^{(i)}}}{\ell_y}
\end{equation}

and the overall average of the $i$-th attribute from all tuples is :

\begin{equation}
\overline{x}^{(i)}=\frac{{\sum_{y \in Y}}{\sum_{j=1}^{\ell_y}{x_{jy}^{(i)}}}}{\ell}=\frac{\sum_{y \in Y}{\ell_y \overline{x}_{y}^{(i)}}}{\ell}
\end{equation}

The square of CR\cite{cor_ratio2} between the i-th attribute of the
dataset and the outcome or the class attribute is given by

\begin{equation}
{Cr^2_i}=\frac{{\sum_{y \in Y}} \ell_y(\overline{x}_y^{(i)}-\overline{x}^{(i)})^2}{\sum_{y\in Y}\sum_{j=1}^{\ell_y}{(x_{jy}^{(i)}-{\overline{x}^{(i)}})^2}}
\end{equation}

If the value of the i-th  attribute of the dataset and value of the outcome are linearly related, then both the Correlation Coefficient and the CR will have same value which is equal to the slope of the dependence.

The CR is able to capture non-linear dependencies in all other cases.

\subsection{Example for Computation of Correlation Ratio}\label{}

\begin{table}[ht]
\captionsetup{justification=centering}
\caption{Dataset Descriptions}\label{tabDes}
\label{tab:6}
\centering
\resizebox{9cm}{!}{

\begin{tabular}{| P{2.5cm} | P{2.5cm} | P{2.5cm} |}
\hline
\textbf{Blood Pressure(BP)} & \textbf{Blood Sugar(BS)}  & \textbf{Age-group}\\
\hline
60 & 100 & teenager \\
\hline
75 & 120 & teenager \\
\hline
70 & 90 & teenager \\
\hline
80 & 125 & teenager \\
\hline
65 & 90 & teenager \\
\hline
80 & 110 & middle-aged \\
\hline
75 & 105 & middle-aged \\
\hline
85 & 123 & middle-aged \\
\hline
72 & 92 & middle-aged \\
\hline
90 & 130 & old \\
\hline
80 & 109 & old \\
\hline
120 & 130 & old \\
\hline
100 & 132 & old \\
\hline
95 & 127 & old \\
\hline
85 & 119 & old \\
\hline
\end{tabular}

}
\end{table}

The computation of CR can be illustrated using the following example where we have considered a dataset (shown in Table \ref{tab:6}) of 15 patients of different age groups - \textit{teenager, middle-aged, old} and the class attribute(Y) for the dataset is Age-group. The labels of attribute Age-group(Y) is denoted as $y$ where $y \in ${\textit{teenager, middle-aged, old}}. Let the i-th  attribute (where i=1) in the dataset is BP. The values of the attribute Blood pressure (BP) for these different sets of patients are given as:

For $y=teenager , S_{teenager}=(x_{jy}^{(i)})$; \hspace{1cm} $j=1,\cdots,5$
\begin{flushleft}
\hspace{4.4cm} $={60,75,70,80,65}$ (BP values of 5 teenager patients)
\end{flushleft}

For $y=middle-aged, S_{middle-aged}=(x_{jy}^{(i)})$; \hspace{0.5cm} $j=1,\cdots,4$ 
\begin{flushleft}
\hspace{2.8cm}$= \{80,75,85,72\}$ (BP values of 4 middle-aged patients) 
\end{flushleft}
                                     
For $y=old , S_{old}=(x_{jy}^{(i)})$;  \hspace{1cm} $j=1,\cdots,6$ 
\begin{flushleft}
\hspace{2.8cm}$= \{90,80,120,100,95,85\}$ (BP values of 6 old patients)\linebreak
\end{flushleft}

  \begin{flushleft}

  For $y=teenager$, average of BP(i-th attribute),
  \begin{flushleft}
   $\overline{x}_{teenager}^{({BP})}=70$
  \end{flushleft}   
   
  For $y=middle-aged$, average of BP(i-th attribute),
  \begin{flushleft}  
   $\overline{x}_{middle-aged}^{({BP})}=78$
   \end{flushleft}
    %\linebreak\linebreak
 For $y=old$, average of BP(i-th attribute),
  \begin{flushleft}  
   $\overline{x}_{old}^{({BP})}= 95$
  \end{flushleft}    
   % \linebreak\linebreak
  Overall average of BP for these three different age-group patients, $\overline{x}^{(BP)}= 82$

  \end{flushleft}

The weighted sum of square of the differences between the average BP of each group of patients
and the overall average is
  \begin{flushleft}
  $=5(70-82)^2 + 4(78-82)^2 + 6(95-82)^2$ 
  \end{flushleft}
  \begin{flushleft}
  $=1798$
  \end{flushleft}
whereas the overall sum of the squares of the differences between
the individual BP and  the overall average BP is:\\
   
    $(60-82)^2 + (75-82)^2 + (70-82)^2 + (80-82)^2 + (65-82)^2 + (80-82)^2 + (75-82)^2 + (85-82)^2 + (72-82)^2 + (90-82)^2 + (80-82)^2 + (120-82)^2 + (100-82)^2 + (95-82)^2 + (85-82)^2 = 3146$

Thus, from equation (7),
   \begin{flushleft}
   ${Cr^2}_{BP} = \frac{1798}{3146} = 0.572$\\
   ${Cr}_{BP} = 0.756$
   \end{flushleft} 
\subsection{Proposed Algorithms}\label{}
The CR approach that has been discussed above is basically applicable to quantitative data.
So, we propose an approach based on the concept of CR which will be applicable on dataset having nominal or categorical attributes.

\begin{algorithm}[H]

\caption{Constructing CR based Decision Tree(D,$N_{\ell}$)}

 \label{algoDecTree}

\begin{algorithmic}[1]
 
\STATE //Inputs: Dataset D and node $N_{\ell}$
\STATE Let $A={A_1, A_2,\cdots, A_n}$ be the set of $n$ attributes for the tuples in D

   \IF {all the tuples in D have the same class label}
     \STATE Return $N_{\ell}$ as a leaf node labelled with the class label

    \ELSE
        \FOR{each attribute $A_i$}
            \STATE $C_{r_i}$ = Correlation\_ratio($A_i,Y$) \hspace{10 mm}, \hspace{1 mm}where $Y$ is the
                   class attribute
                   \STATE Insert $C_{r_i}$ in set CR, \hspace{1 mm}where CR is the set of Correlation             
                          ratios.
        \ENDFOR
        \STATE  $r=max(CR)$
        \IF {multiple $A_i$ have $C_{r_i}==r$}
            \STATE choose the attribute $A_i$ as the splitting attribute which has most of the possible
                   distinct values present in D

        \ELSE
            \STATE Choose the attribute $A_i$ as the splitting attribute which has $C_{r_i}==r$
         \ENDIF

         \STATE Label node $N_{\ell}$ with attribute $A_i$
            \STATE Let attribute $A_i$ has $m$ distinct values $A_{i,1}$,$A_{i,2}$,...,$A_{i,m}$
            
            \STATE Divide $D$ into $m$ partitions $D={D_1,D_2, \ldots,D_m}$ corresponding to each distinct value of $A_i$ respectively, and create a child node $N_{{\ell}j}$ corresponding to each partition from node $N_{\ell}$ with corresponding distinct value of attribute $A_i$ as the label on the branch
            \FOR{each partition $D_j$ in $D$}
                \IF {that partition is empty}
                    \STATE Label node $N_{{\ell}j}$ as a leaf node with the majority class in $D$                       \ELSE
                     \STATE  Call DecisionTree($D_j$,$N_{{\ell}{j}}$)
                \ENDIF
            \ENDFOR

    \ENDIF

\end{algorithmic}

\end{algorithm}

\begin{algorithm}[H]

\caption{Compute CR($A_i$,$Y$) \hspace{20 mm}}

\begin{algorithmic}[1]

\STATE //Inputs: $A_i,Y$
\STATE Let attribute $A_i$ has $m$ distinct values with respect to $D$
\STATE Let  class attribute $Y$ has $l$ distinct labels {$Y_1$,$Y_2$,.....$Y_l$}
\FOR{each class label $Y_j$ of $Y$}
    \FOR{each distinct value $A_{i,k}$ of $A_i$ with class label $Y_j$}
        \STATE $f_k =  frequency(A_{i,k}^{Y_j})$
        \STATE Insert  $f_k$ in set $F_{Y_j}$
    \ENDFOR
    \STATE $m_{Y_j} = max( F_{Y_j} )$
    \STATE ${\bar{x}^i}_{Y_j}=\frac{m_{Y_j}}{t_{Y_j}}$ \hspace{5 mm} , where $t_{Y_j}$ is the total occurrence\hspace{10 mm} of records in the dataset $D$ with class label ${Y_j}$.
\ENDFOR

\FOR{each attribute $A_i$}
\STATE Call  Avg($Y_1$,$Y_2$,....,$Y_l$,$m_{Y_1}$,$m_{Y_2}$,...$m_{Y_l}$,$t_{Y_1}$,$t_{Y_2}$,...,$t_{Y_l}$) \hspace{10 mm} {\raggedleft // This  function returns the value $ \bar{x}^i = \frac{\sum_{Y_j\in Y}m_j}{\sum_{Y_j\in Y}t_{Y_j}} $ \hspace{3 mm} ,where $\bar{x}^i$ is the overall average of the i-th attribute}
\ENDFOR

\STATE Call Disin($Y_1$,$Y_2$,....,$Y_l$,$t_{Y_2}$,...,$t_{Y_l}$,${{\bar{x}}^i}_{Y_1}$, ${{\bar{x}}^i}_{Y_2}$,... ${{\bar{x}}^i}_{Y_l}$,${{\bar{x}}^i}$) \hspace{10 mm} {\raggedleft//which returns the dispersion among individual classes as $ d_{in} = \sum_{Y_j\in Y}t_{Y_j}{({{{\bar{x}}^i}_{Y_j}}- {\bar{x}}^i)}^2 $
\hspace{3 mm} , where ${{{\bar{x}}^i}_{Y_j}}$ is the average of the i-th attribute within each outcome class $Y_j$.}

\STATE Call Disov($Y_1$,$Y_2$,....,$Y_l$,$m$,${x^i}_{(1,{Y_1})}$,${x^i}_{(2,{Y_1})}$,...,${x^i}_{(m,{Y_1})}$,${x^i}_{(1,{Y_2})}$,${x^i}_{(2,{Y_2})}$,...,${x^i}_{(m,{Y_2})}$,...,${x^i}_{(1,{Y_l})}$,${x^i}_{(2,{Y_l})}$,\hspace{10 mm}...,${x^i}_{(m,{Y_l})}$,${{\bar{x}}^i}$) // which returns the dispersion across whole population as
  $d_{ov} = \sum_{Y_j\in Y} \sum_{a=1}^{m}({{x^i}_{(a,{Y_j})}}-\bar{x}^i)^2$, where ${{x^i}_{a,{Y_j}}}$ is the frequency of occurrence of the $a-th$ distinct value of $A_i$ with class label $Y_j$ .
\STATE Compute Correlation Ratio square $ {{Cr}^2}_{A_{i}} $ as the ratio of $d_{in}$ and $d_{ov}$

\STATE Return square-root of ${{Cr}^2}_{A_{i}}$ as Correlation Ratio $Cr_{A_{i}}$

\end{algorithmic}
%}
\label{algoCorrelationratio}
\end{algorithm}

We show in Algorithm \ref{algoDecTree} how to create a Decision
Tree using CR as the splitting criterion. A root
node is built corresponding to the whole dataset. CR based approach is used to split the dataset further. CR between each attribute and the class attribute is calculated at
each level of Decision Tree construction and the attribute with
the highest CR value with the Class attribute is
chosen as the partitioning attribute for the dataset. The root
node is labelled with the corresponding splitting attribute. The
subtrees of the root node are built using the different distinct
values of the splitting attribute as the branch labels and child
nodes are created from the root node for each splitted sub-dataset
respectively. If in any partition, all the tuples have the same
class label, then label the corresponding leaf node with the
corresponding class label. On the other hand, if any partition is
empty, then mark the corresponding leaf node with the majority
class label in its parent's partition. Iterate the same process
until for each partition, all the data points have the same class labels.

In Algorithm \ref{algoCorrelationratio} we have shown various steps to compute CR. When calculating the average value of the i-th attribute within each outcome class we have taken the ratio of the highest frequency value of occurrence for a distinct value of i-th attribute for that class and the total occurrence of records with that outcome class. Thereafter the overall average of the i-th attribute is calculated. Then the ratio of the dispersion among individual classes and the dispersion across the whole population for the i-th attribute is calculated which is actually the square of the CR for the i-th attribute from which the square root is calculated to get the actual CR value.

The computation of significance of an attribute using Algorithm 2 is illustrated using the following example in Table \ref{tab:7}:
 
\begin{table}[H]
\captionsetup{justification=centering}
\caption{Example dataset}
\label{tab:7}
\centering
\resizebox{7cm}{!}{

\begin{tabular}{| P{1cm} | P{1.5cm} | P{1.5cm} | P{1.5cm} |}
\hline
 & \multicolumn{3}{c|}{\textbf{Temperature}}\\ \cline{2-4}
 
 \textbf{Class} & \textit{Hot}  & \textit{Mild} & \textit{Cool}\\
\hline
\textit{No} & 2 & 1 & 0\\
\hline  
\textit{Yes} & 1 & 1 & 2\\
\hline 
\end{tabular}
}
\end{table}

$x_{No}^{(1)}=\frac{2}{3}=0.667$\hspace{2cm}
$x_{Yes}^{(1)}=\frac{2}{4}=0.5$\hspace{2cm}
$x^{(1)}=\frac{4}{7}=0.571$\\
\\
\begin{equation*} 
{Cr^2}_{Temperature}\\
= \frac{3*(0.667-0.571)^2 + 4*(0.5-0.571) ^2}{\parbox{2.5in}{$(2-0.571)^2 + (1-0.571)^2 + (0-0.571)^2 + (1-0.571)^2 + (1-0.571)^2 + (2-0.571)^2$}}\\
\end{equation*}
\hspace{6.8cm}$=0.00963$\\
\parbox{4in}{\hspace{4.55cm}${Cr}_{Temperature}= 0.098$}\\
  In Table \ref{tab:7}, we consider Temperature as the first attribute of a dataset and it has three possible values - \textit{Hot}, \textit{Mild}, \textit{Cool}. There are two classes - \textit{No} and \textit{Yes}. The frequencies of the attribute values for each of the class is shown using the numeric values in each cell of the table. The maximum frequency value for the attribute Temperature is used to calculate the average weight of the attribute in each class.  The overall average weight $x^{(1)}$ of the attribute is the ratio of the summation of the maximum frequencies of the two classes and the total number of instances in the two classes. The significance of the attribute Temperature for predicting class Y, $Cr_{Temperature}$ is calculated as shown in the example. The continuous attributes in the datasets taken from UCI machine learning repository \cite{Lichman:2013} were  discretized. In each level while constructing the Decision Tree, we have considered the attribute which has the highest value of CR with the class attribute.

 \section{Observations on Benchmark Healthcare Datasets}\label{sectabdata}

Several datasets like Pima Indian Diabetes dataset, Liver Disorder dataset, Mammography Masses dataset, Breast Cancer dataset, Hepatitis dataset, Post-operative dataset, ILPD dataset and Spect-heart datasets from UCI machine learning repository were considered for performance evaluation of our proposed approach.

\subsection{About the datasets}\label{subsecDataset}

Table \ref{tab:4} shows the characteristics of the datasets considered here.
\begin{table}[H]
\captionsetup{justification=centering}
\caption{Nature of the Datasets}\label{tabDes}
\label{tab:4}
\centering
\resizebox{17cm}{!}{

\begin{tabular}{| P{3cm} | P{2cm} | P{2cm} | P{6cm} |}
\hline
\textbf{Dataset} & \textbf{Number of instances}  & \textbf{No. of attributes}  & \textbf{Characteristics of attributes}\\
\hline
Pima Indian Diabetes & 768 & 9 & All have same no. of distinct values \\
\hline
Mammography & 961 & 6 & Different no. of distinct values \\
\hline
Breast Cancer & 699 & 10 & All have same no. of distinct values \\
\hline
Hepatitis & 155 & 20 & Different no. of distinct values \\
\hline
Post-operative & 90 & 9 & Different no. of distinct values \\
\hline
ILPD & 583 & 10 & Different no. of distinct values \\
\hline
Spect-heart & 267 & 23 & All have same no. of distinct values \\
\hline
Statlog(heart) & 270 & 13 & Different no. of distinct values \\
\hline
\end{tabular}

}
\end{table}

The Pima Indian Diabetes dataset has overall 9 attributes including the class attribute
which is categorical (\emph{tested positive for diabetes}(1), \emph{tested negative for diabetes}(0)). All the attributes are numeric-valued except the class attribute. There were 768 instances. Among these 500 instances are having class value 0 and 268 instances are having class value 1.

The Mammography Masses dataset consists of 961 instances and 6 attributes in which one of the attribute is the class attribute (possible values - 0 and 1). One attribute was of type integer and the other attributes are nominal or ordinal. The integer attribute was discretized into 5 different values. Two of the ordinal attributes have 5 different values and remaining two ordinal attributes have 4 different values. There were some missing values in almost all of the attributes. Class distribution was like : benign(0)- 516, malignant(1)- 445.

The Breast Cancer dataset has 699 instances and total 10 attributes. The class attribute takes two possibles values - 2 for benign and 4 for malignant. Except the class attribute and another attribute which is indicating the id number, rest of the attributes can take values in the the range 1-10. There are 458 instances which are Benign and 248 Malignant instances. Missing values were present in 16 instances.

The Hepatitis dataset has overall 155 instances and 20 attributes of different types like categorical, integer and real. Six of the attributes are of integer type and were discretized into five discrete values. Thirteen attributes are categorical each having two possible values. The class attribute is categorical and can take two possible values - DIE(32 instances) and LIVE(123 instances). There are missing values in many attributes.

The Post-operative dataset is having 9 attributes (including the class attribute) and 90 instances. Attributes are of different types like - categorical and integer. There are seven categorical attributes and one integer attribute. The class attribute is categorical and can take three possible values - I (patient sent to Intensive Care Unit): 2 instances, S (patient prepared to go home): 24 instances and A (patient sent to general hospital floor): 64 instances. There were no missing values. 

The Indian Liver Patient Dataset(ILPD) contains 583 instances and 10 attributes of different categories like integer and real. Among 10 attributes 9 are integer type and 1 attribute is categorical. Each of the integer type attributes are discretised into five distinct categorical values. The class label attribute can take two possible values '1' and '2'. There are 416 instances with class label '1' and '167' instances with class label '2'.

The Spect Heart dataset consists of 267 instances and 23 attributes including the class attribute. All the attributes are binary. The class attribute can have 2 possible values '0' and '1' and class distribution is - 55 examples with class label '0' and 212 examples with class labels '1'. It has no missing values. 

The Statlog (heart) dataset has 13 attributes where one of the attributes is the class attribute. The attributes are of different types like Real, Ordered, Nominal and Binary  and there are total 270 observations and two classes. Out of six real attributes, five are transformed into five categorical values and the remaining one real attribute is transformed into four categorical values. Three attributes are binary. One attribute is ordered and has three different values. Three attributes are nominal out of which one attribute is having four different values and two attributes is having three different values. 

The continuous attributes of different datasets are discretized. The natures of the datasets after discretization has been shown in Table \ref{tab:4}. For most of the datasets k-fold cross validation has been used where the dataset is divided into k disjoint subsets and (k-1) subsets are used for training and the remaining subset is used for testing. This process is repeated k-number of times and the results of all the iterations are combined together\cite{Han06}. For the Post-Operative dataset, it is divided into training and test sets in the ratio of 70:30. In this case, as the dataset is very small, so cross 
validation is not used. The Spect-heart dataset is already divided into separate training (80 instances) and test sets(187 instances).

\subsection{Result and Analysis} \label{subsecDecTree}
Next we have applied our proposed technique on the datasets discussed in Subsection \ref{subsecDataset}.

\begin{table}[ht]
\captionsetup{justification=centering}
\caption{Experimental Results}\label{tabAttr}
\label{tab:5}
\centering
\resizebox{14cm}{!}{

\begin{tabular}{| P{2cm} | P{3cm} | P{2cm} | P{2cm} |}
\hline
\textbf{Dataset} & \textbf{Cross validation}  &  \textbf{IG Accuracy} &  \textbf{CR Accuracy}\\
\hline
Pima Indian Diabetes & \texttt{5-fold} & 70.93\% & 69.24\%\\
\hline
Mammography & \texttt{5-fold} & 94.11\% & 93.88\%\\
\hline
Breast Cancer & \texttt{5-fold} & 91.03\% & 90.03\%\\
\hline
Hepatitis & \texttt{2-fold} & 71.19\% & 73.78\%\\
\hline
Post-operative & \texttt{70:30} & 62.96\% & 62.96\%\\
\hline
ILPD & \texttt{5-fold} & 66.89\% & 67.57\%\\
\hline
Spect-heart & \texttt{80 training and 187 test instances} & 74.33\% & 74.33\%\\
\hline
Statlog(heart) & \texttt{2-fold} & 74.07\% & 74.4\%\\
\hline
\end{tabular}

}
\end{table}
 
 Table \ref{tab:5} shows that for Pima Indian Diabetes, Mammography and Breast Cancer datasets the IG based approach has performed slightly better than CR based approach. For two of these datasets (Pima Indian Diabetes, Breast Cancer) more number of attributes are there and all of the attributes have same number of distinct values. One dataset (Mammography) has different numbers of distinct values for the attributes but less number of attributes.

 For the Spect-heart dataset both IG and CR approaches have given same performance (74.33\% accuracy). This dataset also has same number of distinct values for all the attributes and has lots of attributes compared to the number of instances. For the Post-operative patient dataset both the approaches have given 62.96\% accuracy. The reason behind the not-so-good performance can be attributed to the fact that there were less number of instances (90), more number of attributes (9) and more number of classes (3) compared to the other datasets. Also this dataset has different number of distinct values for different types of attributes. 
 
 For the ILPD dataset, almost all the attributes have same number of distinct values except one attribute. The CR based approach performed slightly better than the IG based approach. The CR approach outperformed the IG based approach for the Hepatitis dataset. 6 attributes have same number of distinct values and rest 13 attributes have less number of distinct values. The CR approach has given slightly better result for the Statlog (heart) dataset where the attributes have different numbers of distinct values and there are total 270 observations and two classes. 
 
 It is therefore observed from the analysis of the result that generally our proposed approach can handle datasets with same as well as different numbers of distinct valued attributes almost equally well because it is not biased towards attributes with large number of distinct values. On the other hand IG approach prefers attributes with many distinct values. In healthcare datasets where there are large number of attributes and different attributes have different numbers of distinct values, the IG based approach prefers attributes with more number of distinct values rather than attributes which have less number of distinct values even if those attributes may be more significant for classification. This gives a reason for lesser performance of the IG based than our proposed approach in those cases. Also another observation that has been made out of the results is that for smaller datasets where less number of instances and comparatively more number of attributes are there like Hepatitis and Statlog (heart)datasets, the proposed approach performs slightly better than the IG based approach. For this to be taken as a proven fact some more datasets need to be analysed in future.

\section{Conclusion}\label{secConclusion}

Healthcare datasets are available in plenty. The nature or the
distribution patterns of such healthcare datasets may also vary. However we observe that such datasets generally have large number number of attributes and different attributes have different numbers of distinct values. Thus, applying existing IG based splitting criterion may not give good accuracy for
all cases. So, in this paper, CR based Decision Tree learner is proposed. This technique serves as a complement of IG based technique i.e., when IG fails, CR based technique succeeds. We demonstrated this
fact using some benchmark healthcare datasets. In future we would like to explore some more such datasets and apply our proposed technique.

\end{document}